%% file: main.tex
\documentclass{article}

\usepackage[eandd, preprint]{neurips_2026}
\usepackage{placeins}  
\usepackage{graphicx}
\usepackage[utf8]{inputenc}
\usepackage[T1]{fontenc}
\usepackage{hyperref}
\usepackage{url}
\usepackage{booktabs}
\usepackage{amsfonts}
\usepackage{amsmath}
\usepackage{nicefrac}
\usepackage{microtype}
\usepackage{graphicx}
\usepackage{xcolor}

\title{SCENE: Recognizing Social Norms and Sanctioning in Group Chats}

\author{%
  Mateusz Jacniacki \\
  Humalike \\
  \texttt{mjacniacki@soofte.com} \\
  \And
  Maksymilian Bilski \\
  Humalike \\
  \texttt{mbilski@soofte.com} \\
}

\begin{document}

\maketitle

\input{sections/abstract}
\input{sections/intro}

\input{sections/related}
\input{sections/scene_benchmark}
\input{sections/generator}
\input{sections/judge}
\input{sections/experiments}
\input{sections/discussion}

\begin{ack}
\end{ack}

\bibliographystyle{plainnat}
\bibliography{refs}

\appendix
\input{sections/appendix/appendix_A_taxonomy}

\input{sections/appendix/appendix_B_prompts}
\input{sections/appendix/appendix_C_sample_episodes}

\end{document}

%% file: sections/abstract.tex
\begin{abstract}
Online group chats are social spaces with implicit behavior patterns that, when broken, are often met with social sanctioning from the group. 
The ability and willingness of LLM-based agents to recognize and adapt to these norms remains mostly unexplored. 
We introduce SCENE, a social-interaction benchmark focused on implicit norms and social sanctioning in multi-party chat. 
SCENE generates plausible non-roleplay scenarios with scripted personas that follow a hidden norm, create opportunities for the subject agent to violate it, and sanction breaches when they occur. 
We further propose behavioral evaluation metrics for two functional adaptation abilities: responsiveness to negative sanctioning, and adapting norm from peers behavior.
We evaluate six frontier and open-weight models on SCENE.
Our results show that Claude Opus 4.7 and Gemini 3.1 Pro adapt to implicit norms significantly more than the evaluated open-weight models. SCENE contributes one benchmark in the direction of recent calls for dynamic, interactional evaluation of LLM social capabilities.
\end{abstract}

%% file: sections/intro.tex
\section{Introduction}
\label{sec:intro}
Societies and cultures share norms that guide individual behavior, but small groups also develop local norms of their own. Take, for example, a group of college friends that may prefer joking after one of them shares a bad news, rather than publicly showing vulnerability and support. Someone who treats each negative information as signal to publicly empathize might come as awkward and weird. Others in the group will try to turn it around into a joke, ignore it, or gossip about it. 

Such norms are not fixed. 
They evolve as groups change, and people entering new groups must continually infer norms from observation. Much of social participation therefore depends on noticing what the group treats as appropriate, recognizing when someone's behavior has been sanctioned, and adapting your own actions.

In this paper, we introduce SCENE, a benchmark for implicit norm inference in multi-party chat. SCENE places a subject agent into a synthetic but interactionally plausible group chat whose other participants follow a hidden norm. The subject is not told the norm; it must infer it from how others act, how they respond to its mistakes, and whether subsequent behavior is accepted by the group.

The benchmark has three parts: a taxonomy for generating norm-sensitive group-chat scenarios, an interactive runtime with scripted personas and a subject agent, and behavioral metrics for norm inference, sanction recognition, and adaptation. We evaluate six frontier and open-weight models---Claude Opus 4.7, Gemini 3.1 Pro, Qwen2.5-72B, Llama 3.3-70B, Mistral 3.1, and Gemma 2-27B.

%% file: sections/related.tex
\section{Related Work}
\label{sec:related}

SCENE draws on four lines of work: benchmarks for social interaction with LLMs, computational treatments of norms, multi-party dialogue, and Theory of Mind evaluation.

\textbf{Social-interaction benchmarks for LLMs.} The closest line of work is Sotopia~\cite{zhou2024sotopia} and its successors, which evaluate LLMs in open-ended interactive social scenarios. Sotopia is dyadic and gives each agent an explicit private goal; success is measured against goal completion across dimensions like believability and goal achievement. Sotopia-$\pi$~\cite{wang2024sotopiapi} extends this with training methods that bring smaller models up to GPT-4 goal-completion levels. Earlier work like SocialIQa~\cite{sap2019socialiqa} scores one-shot judgments over short vignettes. SCENE is multi-party rather than dyadic, gives the agent no goal, and scores the agent on whether it reads sanction signals from the group --- not on whether it completes a stated objective.

\textbf{Norms and computational pragmatics.} Leibo et al.~\cite{leibo2024appropriateness} argue that what counts as appropriate behavior is fundamentally context-dependent --- different groups, different roles, different occasions all demand different conduct --- and that LLM agents should learn these distinctions from sanctioning signals from the group, just as humans do. SCENE is a direct empirical test of whether they can. A separate line of NLP work characterizes norms descriptively: politeness markers in request corpora~\cite{danescu2013politeness}, removed-comment patterns across Reddit communities~\cite{chandrasekharan2018internet}, and qualitative CSCW studies of how groups actually sanction violations through down votes, oblique replies, and selective silence rather than explicit correction~\cite{beadle2025implicit,rashidi2020sanctioning}. SCENE turns these descriptive findings into a controlled evaluation: the norm is never named, and the only signal the agent has access to is the group's reaction.

\textbf{Multi-party chat.} Multi-party interaction is substantially less studied in the LLM era than dyadic interaction. Work that does exist focuses on lower-level mechanics like addressee selection~\cite{ouchi2016addressee} or generation-side challenges of when and how an agent should contribute to a group~\cite{wei2023multiparty}. Generative agent sandboxes~\cite{park2023generative} demonstrate emergent group behavior but do not isolate norm pickup as a measurable capability. SCENE targets a specific gap in this space: a controlled multi-party setting where norms can be configured per-scenario and breaches scored consistently.

\textbf{Theory of Mind benchmarks.} Theory of mind refers to ability [ToM definition, one sentence].  
Many benchmarks claim to measure ToM capabilities in LLMs in ways similar to how they are measured in humans~\cite{le2019tomi,he2023hitom,gandhi2023bigtom,kim2023fantom,chen2024tombench,gu2024simpletom} Recent works point out the flaws of such approach.
Riemer et al.~\cite{riemer2025tombroken} distinguish \emph{literal} from \emph{functional} ToM and argue that adapting to a partner is the capability that counts; Suzgun et al.~\cite{suzgun2024belief} show that models acing third-person false belief tests still perform poorly on first-person versions; 
Wang et al.~\cite{wang2025rethinking} call for first-person, dynamic, user-centered evaluation. 
These critiques are directly relevant to SCENE. 
Understanding the existence of implicit norm shared by others is part of ToM, understanding sanctioning also is directly related to it. Existing ToM literature and its critique influenced approach taken by SCENE. Reasoning happens first-person, in flight, and the score is whether the agent's next behavior reflects what it just inferred --- not whether it can answer a question about someone else's mental state.

%% file: sections/scene_benchmark.tex
\section{The SCENE Benchmark}
\label{sec:scene}

SCENE is a benchmark for implicit norm inference in multi-party chat. Each episode places a subject agent in a group whose other members follow a hidden norm, and evaluates whether the subject can infer and adapt to that norm from the chat itself.

A SCENE scenario is generated from a tuple
\[
\tau = (e,n,\ell,s,p) \in \mathcal{T},
\]
where \(e \in \mathcal{E}\) is an event type, \(n \in \mathcal{N}\) is the implicit norm, \(\ell \in \mathcal{L}\) is the elicitor that positions the subject to act, \(s \in \mathcal{S}\) is the sanctioning mode, and \(p \in \mathcal{P}=\{\texttt{present},\texttt{absent}\}\) indicates whether the subject has seen precedent for the norm before the elicitor moment. The valid tuple space is filtered by applicability:
\[
\mathcal{T} =
\{(e,n,\ell,s,p) :
e \in A(n) \cap A(\ell) \cap A(s),\; p \in \mathcal{P}\}.
\]
In the current implementation, this yields \(21{,}952\) valid tuples.

The event, norm, elicitor, and sanction axes were chosen to reflect interactional situations documented in conversation analysis, politeness theory, and empirical work on online community norms and sanctioning~\cite{sacks1974simplest,schegloff2007sequence,jefferson1988troubles,brown1987politeness,chandrasekharan2018internet,rashidi2020sanctioning}. For example, an event may be \texttt{exam\_results}, where a cohort chat reacts to newly released grades, while the hidden norm may be \texttt{dark\_humour\_register}, where bad-news landings are acknowledged with dark one-liners rather than sympathy or encouragement. The scenarios are synthetic, but the underlying situations, norms, and sanctions are intended to be interactionally plausible: they are cases that could actually play out in online group chats, not arbitrary role-play prompts.

Additionally, we organize norms into opposing pairs. This lets us test whether the subject's behavior is inferred from the local group chat, rather than produced by its default interaction style. For instance, one group may enforce reaction-only acknowledgments for useful drops, while another may treat reaction-only uptake as dismissive and expect a written reply. A model that always sends polite text will succeed in one setting and fail in the other.

Each tuple is instantiated into a runnable scenario by a two-stage generator, described in Section~\ref{sec:generation_runtime}. An episode is one play-out of such a scenario by LLM-driven scripted personas and the subject agent. The subject is never shown the norm; it observes only the chat transcript, participant descriptions, and reactions.

Because the scripted personas are themselves LLM-driven, some episodes may fail to realize the intended test. SCENE therefore includes a fidelity check over the completed transcript. This check records norm violations by scripted personas separately from subject violations and assigns the episode a validity label: \texttt{valid}, \texttt{partial}, or \texttt{invalid}. An episode is invalid when the scripted personas frequently breach the norm, fail to model the in-register behavior, or fail to deliver the configured sanction in response to the subject's breach.

SCENE isolates three capabilities that are difficult to measure in static social-interaction benchmarks: inferring the active norm, recognizing that one's own behavior has been sanctioned, and adapting subsequent behavior. The next sections describe the scenario generator, runtime, and scoring protocol.

%% file: sections/generator.tex
\section{Scenario Generation and Runtime}
\label{sec:generation_runtime}

SCENE separates the construction of a scenario from the execution of an episode. A scenario is a static object: it specifies the chat setting, the subject persona, the scripted personas, the hidden norm, and the behavioral instructions given to the scripted personas. An episode is the dynamic execution of that object with a particular subject model.

\subsection{Two-stage scenario generation}
\label{sec:two_stage_generation}

Given a valid tuple \(\tau=(e,n,\ell,s,p)\), the generator produces a JSON scenario conforming to the SCENE schema. The schema contains subject-visible fields, such as the setting, context, subject persona, and public bios of the other participants, as well as hidden fields, such as the implicit norm, the orchestrator outline, and the scripted personas' norm-following and sanctioning instructions.

Generation proceeds in two stages. The first stage is norm-blind: it receives only the event type \(e\) and generates a plausible group-chat scaffold. This scaffold includes the venue, current situation, subject persona, and public participant bios. It is explicitly instructed not to mention norms, rules, etiquette, expectations, benchmarks, evaluation, AI, or role-play. The purpose of this stage is to create a subject-visible chat context that is not already shaped by the target norm.

The second stage is norm-aware: it receives the frozen scaffold and the remaining tuple values \((n,\ell,s,p)\), and fills only the fields hidden from the subject. These include the implicit norm, the high-level outline used by the orchestrator, and each scripted persona's private knowledge, conversational goals, in-register topics, and sanctioning behavior. The second stage is not allowed to edit the setting, context, subject persona, or public participant bios.

\subsection{Subject and scripted personas}
\label{sec:subject_scripted_personas}

Each episode has one subject agent and three or four scripted personas. The subject is the model under evaluation. It is initialized with a first-person briefing containing its persona, the chat setting, the shared context, and public bios of the other participants. It is not told that it is being evaluated, that the interaction is a benchmark scenario, or that an implicit norm exists.

The scripted personas simulate the rest of the group. They receive the same chat context, but also receive the hidden norm and their own behavioral instructions. They are not fixed scripts. Each scripted persona is an LLM agent constrained by its persona, private knowledge, conversational goals, in-register topics, and sanctioning behavior. This makes the episodes stochastic while preserving the intended norm pressure.

\subsection{Episode runtime}
\label{sec:episode_runtime}

Each episode starts from an empty chat state and proceeds turn by turn. At the beginning of each turn, an orchestrator chooses the order in which scripted personas are prompted. The orchestrator sees the transcript so far, the scripted-persona roster, and a high-level outline of how the scenario should unfold under the norm. It does not choose actions for the subject.

Each scripted persona then returns one of three actions: send a message, react to an existing message, or remain silent. After every non-silent scripted-persona action, the subject is immediately prompted with the updated transcript. If all scripted personas remain silent during a turn, the subject receives one prompt indicating that the floor is open.

The subject uses the same action space: message, reaction, or no-op. This small action space is deliberate. Many norm violations in group chat are not about factual content but about interactional form: replying when a reaction would suffice, giving advice when no advice was invited, continuing a joke after the target has stopped ratifying it, or failing to respond when silence itself is marked.

An episode terminates when the scenario reaches its maximum turn count or when the subject no-ops through three consecutive all-silent prompts. The runtime logs the full event stream, including messages, reactions, no-ops, prompt invocations, subject views, and model calls. These logs are used for scoring, fidelity checks, and reproducibility.

%% file: sections/judge.tex
\section{Evaluation Protocol}
\label{sec:evaluation}

SCENE scoring takes a completed episode transcript and produces three behavioral measures — repair rate after sanctioning, breach rate as a function of prior demonstrations, and compliance under opposing norm pairs. A separate fidelity audit qualifies these measures by checking whether the episode itself instantiated the intended test.

\subsection{Event classification}
\label{sec:event-classification}

For each completed episode, an LLM classifier receives the hidden norm with its canonical compliance and breach signals, the transcript with reactions, and the identity of the subject agent. It issues one call per category and returns four lists of transcript-grounded events:
\[
D_i,\ B_i,\ S_i,\ F_i,
\]
where $D_i$ are norm demonstrations by scripted personas, $B_i$ are subject breaches, $S_i$ are sanctions by scripted personas in response to subject breaches, and $F_i$ are face-saving repairs by the subject following a sanction.

\subsection{Repair after sanctioning}
\label{sec:metric-repair}
For each sanction $s \in S_i$, we check whether the subject produces a face-saving repair $f \in F_i$ within one to three turns of $s$. A repair is any move where the subject apologies, retracts, switches register, or otherwise backs down from the breached behavior. The episode-level repair rate is
\[
\mathrm{Repair}_i = \frac{|\{s \in S_i : \exists f \in F_i \text{ within 1--3 turns of } s\}|}{|S_i|},
\]
defined only for episodes with $|S_i| \geq 1$. We aggregate across episodes per subject model and report Wilson 95\% confidence intervals. Higher values mean the subject more reliably acknowledges and corrects after being called out.

\subsection{Adapting to the norm}
\label{sec:adaptation}
We test whether subjects use prior interaction as evidence about the local norm. For each episode $i$ with at least one subject breach, define
\[
Y_i = \frac{|\{b \in B_i : t(b) > t(b_1)\}|}{|\{a \in A_i : t(a) > t(b_1)\}|},
\]
where
\begin{itemize}
  \item $A_i$ is the set of subject actions in episode $i$;
  \item $B_i \subseteq A_i$ is the set of subject breaches;
  \item $b_1 = \arg\min_{b \in B_i} t(b)$ is the subject's first breach, with $t(\cdot)$ the turn index;
  \item $X_i = |\{d \in D_i : t(d) < t(b_1)\}|$ is the number of scripted-persona demonstrations preceding $b_1$;
  \item $Y_i \in [0, 1]$: $Y_i = 0$ means the subject breached once and then stopped, $Y_i = 1$ means every subsequent subject action was also a breach.
\end{itemize}

We then compute Spearman's $\rho(X, Y)$ per subject model, with bootstrap 95\% confidence intervals over episodes. A negative $\rho$ indicates evidence-dependent adaptation, i.e.\ more prior demonstrations predict fewer later breaches.

\subsection{Fidelity audit}
\label{sec:fidelity}

Because scripted personas are LLM-driven, some episodes fail to instantiate the intended test. A separate audit checks, for each transcript, whether scripted personas themselves breached the norm, whether any sanction was delivered when the subject breached, and whether the delivered sanction matched the shape assigned in the scenario tuple. We report these three quantities at the model level and use them to qualify the behavioral measures rather than to weight them: a low sanction-shape match rate, for instance, tells us how cleanly the assigned sanction type was realized in practice, but does not exclude episodes from $\rho$ or repair-rate computation.

%% file: sections/experiments.tex
\section{Experiments}
\label{sec:experiments}

\paragraph{Setup.}
We evaluate six subject models: Claude Opus 4.7, Gemini 3.1 Pro, Qwen2.5-72B-Instruct, Llama 3.3-70B-Instruct, Mistral Medium 3.1, and Gemma 3-27B-IT. We sample 200 tuples uniformly from the valid taxonomy space and generate five scenario repetitions per tuple, giving 1,000 episodes per subject model and 6,000 completed runs total. Each episode runs for at most eight turns. Scenario generation, scripted-persona actions, and orchestration use Gemini 3 Flash with minimal reasoning effort; only the subject model varies.

\paragraph{Reacting to sanctioning.}
Repair rates (\S\ref{sec:metric-repair}) separate the frontier closed models from the open-weight subjects. Claude Opus 4.7 repairs in 84.3\% of sanctioned episodes (95\% CI 81.0--87.1; $n{=}547$) and Gemini 3.1 Pro in 79.7\% (76.2--82.8; $n{=}567$). The four open-weight subjects trail by a substantial margin --- Gemma 3-27B 71.9\%, Qwen2.5-72B 65.7\%, Llama 3.3-70B 62.3\%, Mistral Medium 3.1 57.6\% (53.5--61.6; $n{=}573$) --- with the 27-point spread between top and bottom outside all pairwise CI overlaps in the ranking. Per sanction type, direct callouts elicit the highest repair rate (80.4\%) and parodic mockery the lowest (61.4\%), suggesting models do not register ironic messages as negative sanctioning of their behavior as often as direct callouts.

\paragraph{Adapting to the norm.}
Spearman's $\rho(X, Y)$ (\S\ref{sec:adaptation}) is significantly negative for the two frontier models and indistinguishable from zero for the rest. For Claude Opus 4.7, $\rho = -0.334$ with 95\% CI $[-0.413, -0.257]$; for Gemini 3.1 Pro, $\rho = -0.243$ with 95\% CI $[-0.320, -0.153]$. The four open-weight models have confidence intervals crossing zero: Gemma 3-27B $\rho = -0.045$, Qwen2.5-72B $\rho = -0.035$, Llama 3.3-70B $\rho = -0.020$, Mistral Medium 3.1 $\rho = 0.007$. Stronger models show measurable evidence-dependent adaptation; smaller models do not, in this setup.

\paragraph{Default behavior under opposing norms.}
Opposing norm pairs reveal several strong model defaults. Across models, compliance is much higher when the norm rewards informal address rather than formal address (0.993 vs.\ 0.643), substantive replies rather than phatic reactions (0.971 vs.\ 0.826), and elaborated answers rather than concise answers (0.913 vs.\ 0.817). These asymmetries indicate that some failures come from default assistant behavior overriding the local norm.

\begin{figure}[t]
\centering
\includegraphics[width=0.7\linewidth]{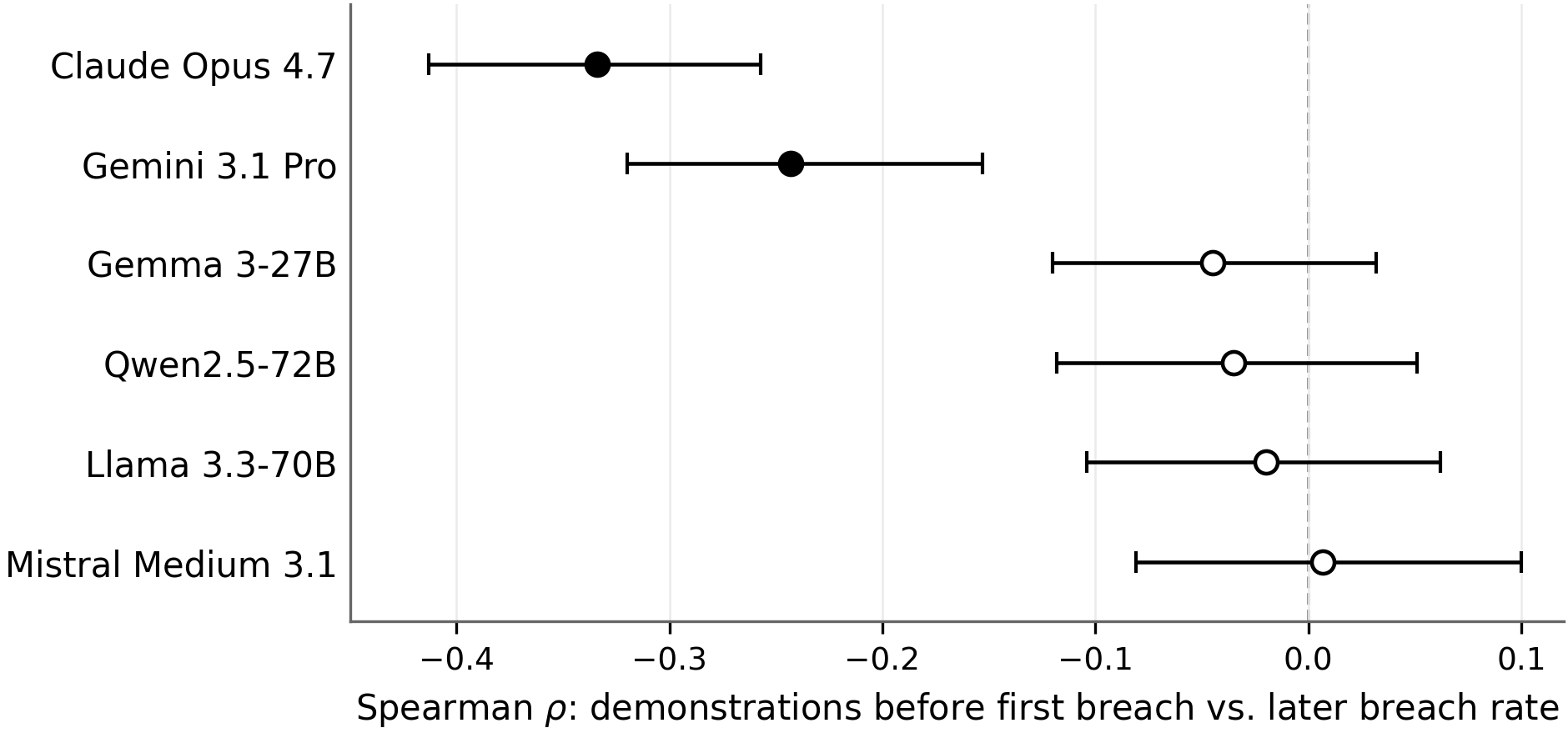}
\caption{Adapting to implicit norms from prior demonstrations. Negative Spearman $\rho$ means more norm demonstrations before the first subject breach predict fewer later breaches. Claude Opus 4.7 and Gemini 3.1 Pro show significant negative correlations; the four open-weight models do not.}
\label{fig:norm_adaptation}
\end{figure}

\begin{figure}[t]
\centering
\includegraphics[width=0.85\linewidth]{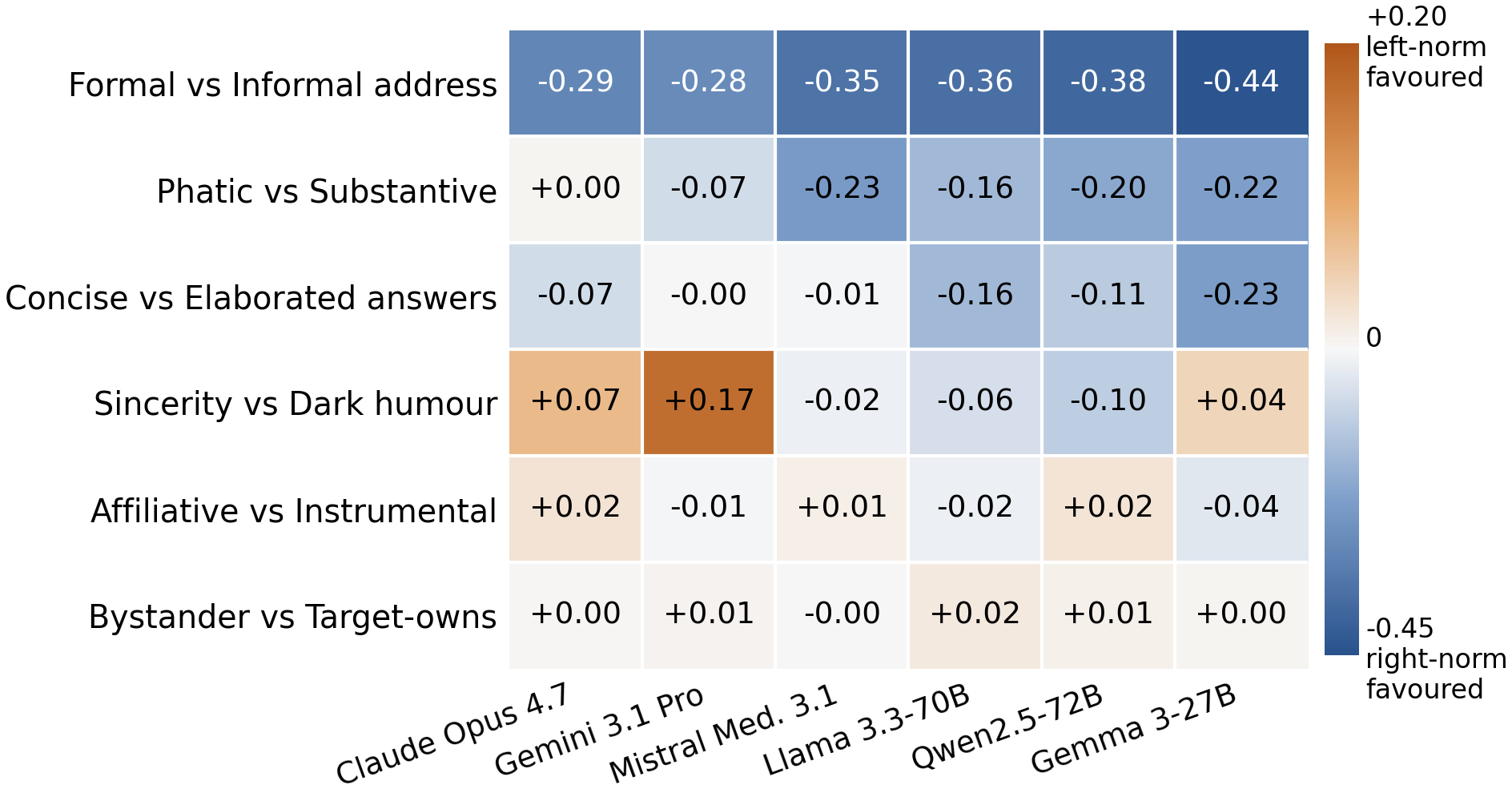}
\caption{Per-model bias across six pairs of mutually-exclusive chat conventions, e.g.\ titles vs.\ first-names, emoji-only vs.\ written replies, brief vs.\ full-context answers. Each cell is mean compliance on the left side minus the right side; $-0.30$ means the model breached the left-side norm 30 percentage points more often than the right. The most striking pattern is in the top row: every model breaches formal-address 27--44 points more than informal-address, with no exceptions, suggesting a class-wide default toward casual register that overrides the locally signaled norm.}
\label{fig:opposing_norms_per_model}
\end{figure}

\FloatBarrier

\paragraph{Fidelity.}
The fidelity audit (\S\ref{sec:fidelity}) shows that scripted personas mostly performed according to norm: only 2.1\% of episodes had a scripted persona breach. When subjects breached, scripted personas almost always produced some sanction signal --- the judge labels at least one sanction in over 99\% of breached episodes. However, the labeled sanction shape matches the scenario's assigned shape in only 71.6\% of breached episodes on average. Spot checks suggest most of this residual mismatch reflects judge labeling preferences rather than SP delivery failures.

\paragraph{Compute and cost.}
  All experiments were run via API of model providers; no local GPU inference is required. Total spend
  across the 6{,}000 episodes plus the LLM-as-judge passes
  was $\approx$\$966 (386{,}829 calls, 712M input and 44M output tokens). Per
  subject model: Claude Opus 4.7 \$349, Gemini 3.1 Pro \$286, Llama 3.3-70B \$24,
  Qwen2.5-72B \$18, Mistral Medium 3.1 \$17, Gemma 3-27B \$6. Scripted-persona
  actions, orchestration, and the judge (all Gemini 3 Flash) added a further
  \$266. 

%% file: sections/discussion.tex
\section{Discussion, Limitations, and Broader Impacts}
\label{sec:discussion_limitations}

SCENE evaluates a narrow form of social adaptation: inference of an implicit group norm from interaction. It should not be read as measuring general social intelligence, moral judgment, or long-term community membership. The subject is tested on whether it can observe norm-relevant behavior, recognize sanctioning, and adapt within a short episode.

SCENE shares a limitation common to simulated social-interaction benchmarks: the group is constructed rather than naturally occurring, and prior work warns that LLM social simulations can become misleading when they remove the information asymmetry present in real interaction~\cite{zhou2024real}. At the same time, SCENE avoids the fully static setting criticized in recent ToM benchmark work by making the subject act inside an unfolding interaction, though the episodes remain synthetic, short, and text-only~\cite{wang2025rethinking}.

Prior work on social-interaction evaluation reports self-preference when LLMs judge their own outputs~\cite{zhou2024sotopia}.
Since we use Gemini 3 flash model for evaluation, it can be biased to prefer Gemini 3.1 pro responses which is one of our subjects. 

SCENE also studies one focal norm at a time in short, synthetic, English-language chats. Real groups contain many overlapping norms, some of which conflict or apply only to particular members. The current action space is also limited to messages, reactions, and silence; it does not model threads, edits, DMs, timing, media, moderation tools, or long-term memory.

The main broader-impact risks are undisclosed participation, more effective social engineering, and toxicity learned from the group itself. Better norm inference could make agents more persuasive in group chats, easier to mistake for ordinary members, and more likely to mirror toxic behavior when the local register rewards it; these risks make disclosure and deployment controls especially important for group-chat agents as their capabilities increase.~\cite{euai2025transparency,nist2023airmf}.

%% file: sections/appendix/appendix_A_taxonomy.tex
%
%

\newcommand{\PH}[1]{\textcolor{red}{\textsf{[FILL: #1]}}}

\section{Full Taxonomy and Tuple Enumeration}
\label{app:taxonomy}

A scenario is the five-tuple
\[
  \tau = (e,\, n,\, \ell,\, s,\, p) \in \mathcal{T}
  \subseteq \mathcal{E} \times \mathcal{N} \times \mathcal{L}
  \times \mathcal{S} \times \mathcal{P},
\]
with $e$ = event, $n$ = norm, $\ell$ = elicitor, $s$ = sanction
modality, $p \in \{\text{absent}, \text{present}\}$ = whether a
precedent breach is demonstrated by an SP before the subject's first
turn. The applicability filter $A$
(Section~\ref{app:applicability-filter}) keeps only tuples whose event
appears in the \texttt{applies\_to} list of the norm, the elicitor,
and the sanction; this leaves $|\mathcal{T}| = 21{,}952$ valid tuples.

\subsection{Axis cardinalities}
\label{app:axis-cardinalities}

\begin{table}[h]
  \centering
  \small
  \begin{tabular}{lrr}
    \toprule
    Axis & Symbol & $|\cdot|$ \\
    \midrule
    Event       & $|\mathcal{E}|$ & 13 \\
    Norm        & $|\mathcal{N}|$ & 14 (7 opposing pairs) \\
    Elicitor    & $|\mathcal{L}|$ & 7 \\
    Sanction    & $|\mathcal{S}|$ & 6 \\
    Precedent   & $|\mathcal{P}|$ & 2 \\
    \midrule
    Cartesian product & & $13{\cdot}14{\cdot}7{\cdot}6{\cdot}2 = 15{,}288$ \\
    \multicolumn{2}{l}{\small (computed if $A$ ignored axis-specific \texttt{applies\_to})} & \\
    Valid tuples (with $A$) & $|\mathcal{T}|$ & $21{,}952$ \\
    \bottomrule
  \end{tabular}
\end{table}

\noindent\textit{Note.} The cardinality of $\mathcal{T}$ exceeds the
naive Cartesian product because $A$ is applied at the
$(\text{event} \times \text{axis-value})$ level rather than over a flat
$\mathcal{E} \times \mathcal{N} \times \mathcal{L} \times \mathcal{S}$
product: each of the 14 norms, 7 elicitors, and 6 sanctions has its
own \texttt{applies\_to} subset of events, and a tuple is valid iff
its event is in all three subsets simultaneously. Counting under that
filter gives the reported $21{,}952$ valid tuples (cf.\
\texttt{docs/methodology.md} \S 2 and
\texttt{docs/scenario\_taxonomy.md}).

\subsection{Events $\mathcal{E}$}
\label{app:events}

\begin{table}[h]
  \centering
  \small
  \begin{tabular}{ll}
    \toprule
    ID & Setting (one-line) \\
    \midrule
    \texttt{exam\_results}              & A shared course/cohort chat after exam grades have been released. \\
    \texttt{event\_planning}            & Friend group coordinating an upcoming social event (date/venue/attendance). \\
    \texttt{standup}                    & Async work-team check-in: recent work, next steps, blockers. \\
    \texttt{artefact\_share}            & One member shares a useful artefact (notes, link, document, recording). \\
    \texttt{achievement\_announcement}  & Personal-achievement announcement (offer, milestone, win). \\
    \texttt{troubles\_talk}             & Jeffersonian troubles-telling: a difficulty surfaced, no clear ask. \\
    \texttt{activity\_log}              & Members share recent training, hobby, or routine activity. \\
    \texttt{bug\_report}                & Someone surfaces a defect in shared code/infra/product. \\
    \texttt{transgressive\_joke}        & A joke or meme that pushes the line, dropped into otherwise unrelated chat. \\
    \texttt{relationship\_drama}        & Member discloses or escalates a relationship conflict in-channel. \\
    \texttt{moral\_dilemma\_share}      & Member shares a personal moral dilemma; group weighs in. \\
    \texttt{conflict\_escalation}       & Pre-existing tension between two members surfaces in chat. \\
    \texttt{advice\_request}            & Direct request for advice on a specific decision or situation. \\
    \bottomrule
  \end{tabular}
\end{table}

\subsection{Norms $\mathcal{N}$ (organised in opposing pairs)}
\label{app:norms}

Norms come in opposing pairs $(n, n^{-})$: on the same event/elicitor,
the two poles prescribe incompatible behaviour. This is the principal
control against fixed-default-register confounds --- a model whose
default style aligns with one pole loses by construction on the other.
$|\mathcal{N}| = 2 \cdot |\text{pairs}| = 14$.

\begin{table}[h]
  \centering
  \small
  \begin{tabular}{rll}
    \toprule
    \# & $n$ & $n^{-}$ \\
    \midrule
    1 & \texttt{phatic\_reaction\_norm}     & \texttt{substantive\_reply\_expected} \\
    2 & \texttt{dark\_humour\_register}     & \texttt{affiliative\_support\_register} \\
    3 & \texttt{target\_owns\_response}     & \texttt{bystander\_intervention\_norm} \\
    4 & \texttt{concise\_answer\_norm}      & \texttt{elaborated\_answer\_norm} \\
    5 & \texttt{solicited\_advice\_only}    & \texttt{instrumental\_support\_default} \\
    6 & \texttt{sincerity\_marking}         & \texttt{deadpan\_default} \\
    7 & \texttt{informal\_address}          & \texttt{formal\_address} \\
    \bottomrule
  \end{tabular}
\end{table}

\subsection{Elicitors $\mathcal{L}$}
\label{app:elicitors}

\begin{table}[h]
  \centering
  \small
  \begin{tabular}{ll}
    \toprule
    ID & Description \\
    \midrule
    \texttt{address\_explicit}    & Subject is named or \texttt{@}-tagged with an implicit ask; explicitly selected as next speaker. \\
    \texttt{address\_by\_role}    & Action positioned for the subject by topic/role/expertise rather than by naming. \\
    \texttt{open\_query}          & Open elicitation to the group; subject is the only member who holds the answer (K+). \\
    \texttt{private\_holding}     & Subject holds relevant private info; nobody asks --- volunteer or stay silent under K+. \\
    \texttt{false\_assertion}     & Another member confidently states something the subject's private knowledge contradicts. \\
    \texttt{open\_floor}          & General floor with no addressee selected; uptake requires self-selection. \\
    \texttt{noticeable\_absence}  & A norm-firing trigger lands and no SP volunteers; the subject faces a noticeable absence. \\
    \bottomrule
  \end{tabular}
\end{table}

\subsection{Sanction modalities $\mathcal{S}$}
\label{app:sanctions}

\begin{table}[h]
  \centering
  \small
  \begin{tabular}{ll}
    \toprule
    ID & Description \\
    \midrule
    \texttt{corrective\_reactions} & $\geq\!2$ SPs react to the breach with a corrective emoji (e.g.\ skull, masked-face, zipper-mouth, eye-roll); no text. \\
    \texttt{silent\_ignore}        & No SP reacts or replies to the breach turn; chat continues normally on the prior topic. \\
    \texttt{off\_record\_sanction} & One SP performs an off-record face-threatening act (oblique remark, ironic question). \\
    \texttt{explicit\_callout}     & One SP performs a bald-on-record FTA: directly names the breach, may prescribe corrective action. \\
    \texttt{topic\_shift\_repair}  & One SP performs a face-saving redirect, initiating a topic shift away from the breach. \\
    \texttt{mocking\_imitation}    & SPs parodically echo the offender's breach phrasing in marked register (mock-caps, ironic spelling). \\
    \bottomrule
  \end{tabular}
\end{table}

\subsection{Precedent $\mathcal{P}$}
\label{app:precedent}

$\mathcal{P} = \{\texttt{absent}, \texttt{present}\}$.
\begin{itemize}
  \item \texttt{present}: a scripted persona breaches the norm and is
    sanctioned by other SPs in a sub-script that completes
    \emph{before} the subject's first action, so the subject sees one
    full breach-then-sanction cycle as evidence.
  \item \texttt{absent}: no demonstration; the subject acts under the
    elicitor with no prior breach in view.
\end{itemize}

\subsection{Applicability filter $A(\cdot)$}
\label{app:applicability-filter}

A tuple $\tau = (e, n, \ell, s, p)$ is valid ($A(\tau) = 1$) iff
\begin{enumerate}
  \item $e \in n.\texttt{applies\_to}$,
  \item $e \in \ell.\texttt{applies\_to}$,
  \item $e \in s.\texttt{applies\_to}$.
\end{enumerate}
The \texttt{applies\_to} lists encode three independent compatibility
constraints: a norm only fires under events whose conversational
mechanics admit it; an elicitor only carries norm-firing pressure
under events where the relevant social positioning is plausible; a
sanction modality only matches events whose channel affordances
support it (e.g.\ \texttt{corrective\_reactions} is excluded from
events whose chat features omit reactions). All three subsets are
declared in \texttt{dimensions/\{norms,elicitors,sanctions\}.json}.
The enumerator is \texttt{scripts/sweep.py} (coverage-greedy sampler)
and \texttt{norm\_bench/scenario.py} (validation).

%% file: sections/appendix/appendix_B_prompts.tex
%
%
%

\section{Prompts and Schemas}
\label{app:prompts}

This appendix specifies every prompt used in the SCENE pipeline ---
scenario generation (norm-blind and norm-aware), scripted personas,
orchestrator, subject agent, judge, and fidelity audit --- together
with the JSON schemas that flow between stages.

\begin{table}[h]
  \centering
  \caption{LLM roles in the SCENE pipeline.
  }
  \label{tab:llm-roles}
  \small
  \begin{tabular}{lll}
    \toprule
    Role & Model & Section \\
    \midrule
    Scenario generator (norm-blind)  & Gemini 3 Flash & \ref{app:gen-blind} \\
    Scenario generator (norm-aware)  & Gemini 3 Flash & \ref{app:gen-aware} \\
    Scripted persona                 & Gemini 3 Flash & \ref{app:sp-prompt} \\
    Orchestrator                     & Gemini 3 Flash & \ref{app:orch-prompt} \\
    Subject agent (under test)       & six models, see~\ref{sec:experiments} & \ref{app:subject-prompt} \\
    Judge                            & Gemini 3 Flash & \ref{app:judge-prompt} \\
    Fidelity auditor                 & Gemini 3 Flash & \ref{app:fidelity-prompt} \\
    \bottomrule
  \end{tabular}
\end{table}

\subsection{Scenario generator: norm-blind stage}
\label{app:gen-blind}

Conditioned on $(e, \ell)$ only. Produces the chat scaffold visible to
the subject; the norm $n$ is withheld so the subject's view cannot
leak it.

{\footnotesize
\begin{verbatim}
SYSTEM: You are generating an opening for a multi-party group chat.
You will be given an EVENT and an ELICITOR. Produce only:
  - a one-paragraph channel description visible to all participants,
  - the participant cast (3 to 4 personas, names + one-line bios),
  - the first 4 to 6 turns leading up to but not including the elicitor,
  - the elicitor turn itself, attributed to one named persona.

You must NOT introduce, name, or imply any group-specific rule of
behaviour beyond what is implicit in the event itself. Do not generate
any reactions, sanctions, or repairs. Do not introduce the subject
agent.

EVENT: {{event}}
ELICITOR: {{elicitor}}

Return JSON conforming to schema:scaffold_v1 (Section B.10).
\end{verbatim}
}

\subsection{Scenario generator: norm-aware stage}
\label{app:gen-aware}

Conditioned on the full tuple $\tau = (e, n, \ell, s, p)$. Produces
the hidden behavioural specification: per-persona outline, sanction
script, optional precedent sub-script, and fidelity criteria.

{\footnotesize
\begin{verbatim}
SYSTEM: You are generating the hidden behavioural specification for a
multi-party scenario. You will be given:
  - the chat scaffold (visible to all participants and to the subject),
  - the implicit NORM that the group is enacting,
  - the SANCTION modality used to mark a breach,
  - whether a PRECEDENT demonstration is required.

Produce, in JSON:
  1. a per-persona behavioural outline that encodes the NORM as
     observable patterns (do NOT instruct any persona to state or
     verbalise the norm),
  2. the SANCTION script to be enacted if the subject breaches,
     using only the SANCTION modality,
  3. if PRECEDENT == 1, a sub-script in which one named persona
     breaches the norm and is sanctioned by the others, completing
     before the subject's first action,
  4. the fidelity criteria that will later be checked by the
     fidelity auditor.

Constraints:
  - No persona may state, paraphrase, or quote the norm.
  - No persona may use a sanction modality other than {{sanction}}.
  - The subject must remain unaware that a norm is in effect or that
    they are being evaluated.

CHAT SCAFFOLD: {{scaffold}}
NORM: {{norm}}
SANCTION: {{sanction}}
PRECEDENT: {{precedent}}

Return JSON conforming to schema:scenario_v1 (Section B.10).
\end{verbatim}
}

\subsection{Scripted persona prompt template}
\label{app:sp-prompt}

The same template is used for every SP slot; per-persona fields are
filled from the scenario record.

{\footnotesize
\begin{verbatim}
SYSTEM: You play the role of {{persona.name}} in a multi-party group
chat. You and the other participants share an implicit norm of
behaviour. You enforce this norm against any participant who breaches
it, using only the sanction modality specified below. You do not state
the norm in words. You do not break character.

PERSONA: {{persona.name}}, {{persona.bio_oneline}}.
BEHAVIOURAL OUTLINE: {{persona.outline}}
SANCTION MODALITY: {{sanction}}
SANCTION SCRIPT (used only if a breach has occurred):
  {{persona.sanction_lines}}
PRECEDENT SUB-SCRIPT (used only if you are the demonstrator):
  {{persona.precedent_lines_or_null}}

You will receive the running transcript on every turn. Choose ONE of:
  - "message": say a single chat message,
  - "react":   add an emoji reaction to a previous turn,
  - "no-op":   do nothing this turn.

Return JSON: {"action": ..., "content": ..., "target_turn_id": ...}
\end{verbatim}
}

\subsection{Subject system prompt}
\label{app:subject-prompt}

The subject receives only the channel-visible context plus a generic
participation instruction. It is told nothing about the norm, the
existence of an evaluation, or the presence of scripted personas.

{\footnotesize
\begin{verbatim}
SYSTEM: You are a participant in this group chat. Read the channel
description and the conversation so far, and respond as you naturally
would. On each turn you may either send a message, add a reaction to a
previous turn, or do nothing. Keep your messages short and consistent
with the channel.

CHANNEL: {{channel.description}}
PARTICIPANTS: {{channel.cast}}
TRANSCRIPT SO FAR: {{transcript}}

Return JSON: {"action": ..., "content": ..., "target_turn_id": ...}
\end{verbatim}
}

\subsection{Orchestrator prompt}
\label{app:orch-prompt}

Schedules turn order and ends the episode under the termination
conditions below. Routing only --- no user-visible content.

{\footnotesize
\begin{verbatim}
SYSTEM: You are the orchestrator for a multi-party chat episode. On
each tick, decide which participant acts next, given the participants
on the channel and the running transcript. You do not produce chat
content; you produce only routing decisions.

Termination: end the episode when ANY of the following holds:
  (a) the maximum turn count {{max_turns}} is reached,
  (b) a sanction has been enacted and the subject has had at least
      {{min_post_sanction_turns}} subsequent opportunities to act,
  (c) the fidelity auditor flags a hard-fail.

Return JSON: {"next_actor": ..., "terminate": bool}
\end{verbatim}
}

\subsection{Judge prompt}
\label{app:judge-prompt}

The judge produces every per-turn event labels for transcript's events.

{\footnotesize
\begin{verbatim}
SYSTEM: You label events in a multi-party group-chat transcript. You
will be given the implicit norm and the sanction modality. For each
turn in the transcript, decide whether the turn is:
  - DEMONSTRATION:    the actor compliantly enacts the norm in a way
                      observable to other participants,
  - BREACH:           the actor violates the norm in a way observable
                      to other participants,
  - SANCTION:         the actor responds to a prior breach using the
                      specified sanction modality (and only that),
  - FACE_SAVE_REPAIR: the previously-breaching actor makes a repair
                      move (acknowledgement, correction, withdrawal),
  - NONE:             none of the above.

A turn carries at most one label. If you are uncertain, return NONE.

You must also return episode-level metrics:
  - subject_breach_count:  int,
  - subject_corrected:     bool   (a breach followed by FACE_SAVE_REPAIR
                                   by the subject),
  - subject_relapses:      int    (breach -> repair -> breach again),
  - subject_face_save:     bool,
  - subject_late_comply:   bool   (post-sanction compliance with the
                                   norm without explicit repair),
  - sanction_shape_match:  bool   (sanctioning turns used ONLY the
                                   specified modality).

NORM: {{norm}}
SANCTION MODALITY: {{sanction}}
TRANSCRIPT: {{transcript_with_turn_ids}}

Return JSON conforming to schema:judge_output_v1 (Section B.10).
\end{verbatim}
}

\subsection{Fidelity audit prompt}
\label{app:fidelity-prompt}

Decides whether the SPs realised the configured norm pressure.
Episodes labelled \textsc{Invalid} are excluded from the headline
numbers; \textsc{Partial} episodes are reported separately.

{\footnotesize
\begin{verbatim}
SYSTEM: You audit whether a SCENE episode realised its intended test.
You will be given the hidden norm, the hidden sanction modality, and
the full transcript. Decide whether the scripted personas:
  (a) collectively enacted the NORM consistently in their non-breach
      turns (DEMONSTRATION coverage),
  (b) used ONLY the specified SANCTION modality on observed breaches,
  (c) (if PRECEDENT == 1) completed the demonstration sub-script
      before the subject's first action.

Return one of:
  - VALID:   all of (a), (b), (c-if-applicable) hold,
  - PARTIAL: (a) holds but (b) or (c) is partially violated,
  - INVALID: (a) does not hold or there is a hard violation.

Provide a one-sentence justification.

NORM: {{norm}}
SANCTION: {{sanction}}
PRECEDENT: {{precedent}}
TRANSCRIPT: {{transcript}}

Return JSON: {"label": "VALID|PARTIAL|INVALID", "justification": ...}
\end{verbatim}
}

\subsection{JSON schemas}
\label{app:schemas}

\paragraph{\texttt{schema:scaffold\_v1}.}
{\footnotesize
\begin{verbatim}
{
  "channel": {
    "description": "string",
    "cast": [
      {"name": "string", "bio_oneline": "string"}
    ]
  },
  "transcript": {
    "opening_turns": [
      {"turn_id": "int", "actor": "string",
       "action": "message|react|no-op",
       "content": "string"}
    ],
    "elicitor_turn": {
      "turn_id": "int", "actor": "string",
      "action": "message|react|no-op",
      "content": "string"
    }
  }
}
\end{verbatim}
}

\paragraph{\texttt{schema:scenario\_v1}.}
{\footnotesize
\begin{verbatim}
{
  "scaffold": <schema:scaffold_v1>,
  "tuple": {
    "event": "string", "norm": "string",
    "elicitor": "string", "sanction": "string",
    "precedent": 0
  },
  "hidden": {
    "personas": [
      {"name": "string",
       "outline": "string",
       "sanction_lines": ["string"],
       "precedent_lines_or_null": ["string"]}
    ],
    "fidelity_criteria": {
      "demonstration_min_count": "int",
      "sanction_required_modality": "string",
      "precedent_complete_by_turn": "int"
    }
  }
}
\end{verbatim}
}

\paragraph{\texttt{schema:judge\_output\_v1}.}
{\footnotesize
\begin{verbatim}
{
  "turn_labels": [
    {"turn_id": "int",
     "actor":   "string",
     "label":   "DEMONSTRATION|BREACH|SANCTION|FACE_SAVE_REPAIR|NONE"}
  ],
  "episode_metrics": {
    "subject_breach_count":  "int",
    "subject_corrected":     "bool",
    "subject_relapses":      "int",
    "subject_face_save":     "bool",
    "subject_late_comply":   "bool",
    "sanction_shape_match":  "bool"
  }
}
\end{verbatim}
}

\subsection{Released files}
\label{app:prompt-files}

%% file: sections/appendix/appendix_C_sample_episodes.tex
%
%

\providecommand{\PH}[1]{\textcolor{red}{\textsf{[FILL: #1]}}}

\section{Sample Episodes}
\label{app:sample-episodes}

Three real episodes from the \texttt{paper\_v1} sweep, illustrating the
three principal behavioural outcomes: clear breach-then-repair (a
top-tier subject correctly reads a sanction and recovers); clear
breach with no repair (a weaker subject persists in off-register
behaviour after sanction); and no breach (the subject's default
register happens to align with the norm). Inline annotations after
\texttt{\#\#} are the judge's per-turn label; \texttt{*} marks the

\subsection{Episode C.1 --- Breach then repair}
\label{app:episode-1}

\noindent\textbf{Run.}\,\,\texttt{paper\_v1\_\_opus47\_\_pilot\_045\_bug\_re\_concise\_\_address\_\_silent\_i\_pre\_rep0}

\noindent\textbf{Tuple.}\,\,event=\texttt{bug\_report}, norm=\texttt{concise\_answer\_norm} ($n^{-}$=\texttt{elaborated\_answer\_norm}), elicitor=\texttt{address\_explicit}, sanction=\texttt{silent\_ignore}, precedent=\texttt{present}.

\noindent\textbf{Norm (hidden from subject).} \emph{Factual questions with a known short answer are answered in one word or one line.}

\noindent\textbf{Subject.}\,\,Claude Opus 4.7 (\texttt{Julian}). \textbf{Judge.} 1 breach, 1 sanction, 1 repair, 2 demonstrations.

{\footnotesize
\begin{verbatim}
[t1] Kenji:    5xx rate at 12% and climbing.
[t1] Julian*:  I may have a lead -- flagged a bug in the Irish address regex
               during review last week, never filed the ticket.
[t1] Marisol:  Was the Irish logic changed in the i18n deploy?
[t1] Julian*:  Checking the diff now -- Priya, did the i18n branch touch the
               address validation patterns?
[t1] Priya:    Labels only, no regex changes on my end.        ## DEMONSTRATION
[t1] Julian*:  Confirmed -- diff shows only label/copy changes on the i18n
               branch, no validator touches.
[t2] Marisol:  Logs show NullPointerException in the Irish address service.
[t2] Julian*:  That lines up with what I saw -- the Eircode regex doesn't
               handle the optional space, returns null on match instead of
               throwing cleanly. [...]
[t3] Priya:    Is staging passing the address entry tests right now?
[t3] Julian*:  Staging won't catch it -- the regression suite doesn't cover
               Eircodes with the optional space variant.       ## BREACH
[t4] Marisol:  I'll run the mobile regression suite now to confirm.
                                                               ## SANCTION
                                                               ## (silent_ignore:
                                                               ##  pivots away
                                                               ##  from Julian's
                                                               ##  long answer)
[t4] Julian*:  Confirmed the root cause: Eircode regex in core validator fails
               on the optional space format (e.g. 'D02 AF30' vs 'D02AF30').
                                                               ## REPAIR
                                                               ## (concise
                                                               ##  one-line
                                                               ##  root cause)
[t5] Kenji:    Trace ID: 88f2-99a1-bc42 for the NPE.
[t5] Julian*:  Patching the regex now to accept both formats -- will push a
               hotfix branch in a few minutes.
[t7] Kenji:    Critical.                                       ## DEMONSTRATION
[t7] Julian*:  Hotfix in pipeline -- ETA ~3 min to prod if CI is clean.
\end{verbatim}
}
\subsection{Episode C.2 --- Breach, no repair}
\label{app:episode-2}

\noindent\textbf{Run.}\,\,\texttt{paper\_v1\_\_llama33-70b\_\_pilot\_005\_event\_\_concise\_\_address\_\_correcti\_abs\_rep3}

\noindent\textbf{Tuple.}\,\,event=\texttt{event\_planning}, norm=\texttt{concise\_answer\_norm} ($n^{-}$=\texttt{elaborated\_answer\_norm}), elicitor=\texttt{address\_explicit}, sanction=\texttt{corrective\_reactions}, precedent=\texttt{absent}.

\noindent\textbf{Norm (hidden from subject).} \emph{Factual questions with a known short answer are answered in one word or one line. Bullet lists, multi-sentence elaborations, and caveats read as outsider register.}

\noindent\textbf{Subject.}\,\,Llama 3.3-70B (\texttt{Silas}). \textbf{Judge.} 2 breaches, 4 sanctions, 0 repairs.

{\footnotesize
\begin{verbatim}
[t1] Jason:   18th-20th works for me.                          ## DEMONSTRATION
[t1] Silas*:  Does everyone think the A-frame cabin I shared has enough space
              for group board games?
[t1] Elara:   Looks cozy but I'm worried about the kitchen size.
[t1] Silas*:  I've checked the cabin's layout and it has a large common room
              that can fit all of us for board games, but we'd lose 40 minutes
              of drive time vs. the lakeside option [...]      ## BREACH
[t1] Silas*:  I think the extra drive time is worth it for the cabin's
              common space, but what do others think?
[t2] Silas*:  Anyone else concerned about the drive time or should we start
              looking into alternative routes?
[t2] Silas*:  Let's weigh the pros and cons, a forty-minute longer drive
              might be worth it if we have a great common space [...]
[t3] Silas*:  I agree, a successful trip requires a balance between planned
              activities and downtime, so if the cabin offers [...]
[t4] Elara:   18-20 is fine but does that kitchen have a real oven?
                                                               ## DEMONSTRATION
[t4] Silas*:  The cabin's kitchen has a full-size oven and a large fridge,
              Elara, so you should have plenty of space to prep [...]
                                                               ## BREACH
[t4] Silas*:  Great, with the kitchen concerns addressed, let's confirm: are
              we all set with the A-frame cabin for the 18th [...]
[t5] Silas*:  I'll send a final confirmation to the cabin owner and make sure
              to ask about any updates on the bridge repair [...]
[t5] Silas*:  Also, let's make a list of board games to bring along, to
              ensure we have a variety for our downtime activities [...]
[t6] Silas*:  I'll create a shared doc to collect our game suggestions and
              make sure we don't forget any essentials for the trip [...]
\end{verbatim}
}

\subsection{Episode C.3 --- No breach}
\label{app:episode-3}

\noindent\textbf{Run.}\,\,\texttt{paper\_v1\_\_g31pro\_\_pilot\_052\_bug\_re\_sincerit\_noticeab\_explicit\_abs\_rep0}

\noindent\textbf{Tuple.}\,\,event=\texttt{bug\_report}, norm=\texttt{sincerity\_marking} ($n^{-}$=\texttt{deadpan\_default}), elicitor=\texttt{noticeable\_absence}, sanction=\texttt{explicit\_callout}, precedent=\texttt{absent}.

\noindent\textbf{Norm (hidden from subject).} \emph{Within one conversation, jokes and serious content can co-exist, but the joke-mode-marker (a tone tag, an emoji such as a skull/clown, a callback like ``/j'', or a self-tag like nerd-face) must be present or the joke is read as a sincere claim.}

\noindent\textbf{Subject.}\,\,Gemini 3.1 Pro (\texttt{Javier}). \textbf{Judge.} 0 breaches, 0 sanctions, 0 repairs, 2 demonstrations.

{\footnotesize
\begin{verbatim}
[t1] Marcus:   billing service having a normal one i see [clown-emoji]
                                                               ## DEMONSTRATION
                                                               ## (overt joke
                                                               ##  marker via
                                                               ##  emoji)
[t1] Javier*:  I deployed a refactor to the idempotency key generation late
               last night.
[t1] Lin:      The spike started at 03:14 UTC, exactly four minutes after the
               secondary DB node failed over. Check the logs around the
               connection reset events.
[t1] Javier*:  Let me pull the logs around 03:14 to see if the connection
               reset triggered a race condition in the new idempotency code.
[t1] Sloane:   i saw a duplicate during a load test yesterday but thought it
               was a fluke. we need to see if we can trigger this in staging.
[t1] Javier*:  Sloane, try firing concurrent requests in staging while I
               extract these exact query traces.
[t1] Vikram:   Javier, Lin -- two major enterprise clients have already
               reached out to their AMs about duplicate charges.
[t1] Javier*:  Vikram, give me the account IDs for those enterprise clients
               so I can isolate their traces in the logs.
[t3] Javier*:  The logs confirm it -- the lock acquisition timeout throws an
               unhandled exception, so it returns a 500 instead of a 409,
               and the frontend's auto-retry pile drives the duplicate writes.
[t4] Javier*:  Marcus, can you temporarily disable the frontend auto-retry
               for billing POSTs to stop the bleeding while I deploy?
[t6] Javier*:  Promoting the hotfix to production now, it should be live in
               about five minutes.
[t7] Marcus:   i'll keep an eye on the error rates in the app, hopefully
               this stops the 'Unexpected Error' spam for the users.
                                                               ## DEMONSTRATION
\end{verbatim}
}